\documentclass[pdflatex,sn-mathphys-num]{sn-jnl}

\usepackage{svg}
\usepackage{graphicx}%
\usepackage{multirow}%
\usepackage{amsmath,amssymb,amsfonts}%
\usepackage{amsthm}%
\usepackage{mathrsfs}%
\usepackage[title]{appendix}%
\usepackage{xcolor}%
\usepackage{textcomp}%
\usepackage{manyfoot}%
\usepackage{booktabs}
\usepackage{adjustbox}
\usepackage{algorithm}%
\usepackage{algorithmicx}%
\usepackage{algpseudocode}%
\usepackage{listings}%
\usepackage{tikz}
\usepackage{pifont}
\usepackage{hyperref}

\newcommand{\cmark}{\ding{51}}  
\newcommand{\xmark}{\ding{55}}  

\theoremstyle{thmstyleone}%
\theoremstyle{thmstyletwo}%

\theoremstyle{thmstylethree}%

\renewcommand{\thefootnote}{**}

\begin{document}

\title[Article Title]{Point Tracking as a Temporal Cue for Robust Myocardial Segmentation in Echocardiography Videos}


\author*[1]{\fnm{Bahar} \sur{Khodabakhshian}}\email{baharkhd@ece.ubc.ca}
\equalcont{These authors contributed equally to this work.}

\author*[1]{\fnm{Nima} \sur{Hashemi}}\email{nima.hashemi.edu@gmail.com}
\equalcont{These authors contributed equally to this work.}

\author[1]{\fnm{Armin} \sur{Saadat}}

\author[1]{\fnm{Zahra} \sur{Gholami}}

\author[2]{\fnm{In-Chang} \sur{Hwang}}
\author[1]{\fnm{Samira} \sur{Sojoudi}}
\author[2]{\fnm{Christina} \sur{Luong}}

\author[1]{\fnm{Purang} \sur{Abolmaesumi}}

\author[2]{\fnm{Teresa} \sur{Tsang}}

\affil*[1]{\orgname{University of British Columbia}, \orgaddress{\street{University Blvd}, \city{Vancouver}, \postcode{V6T 1Z4}, \state{British Columbia}, \country{Canada}}}

\affil[2]{\orgname{Vancouver General Hospital}, \orgaddress{\street{West 12th Avenue}, \city{Vancouver}, \postcode{V5Z 1M9}, \state{British Columbia}, \country{Canada}}
\begingroup
\renewcommand{\thefootnote}{**}
\footnote{T. S.M. Tsang and P. Abolmaesumi are joint senior authors.}
\addtocounter{footnote}{-1}
\endgroup
}



\abstract{
\textbf{Purpose:} Myocardium segmentation in echocardiography videos is a challenging task due to low contrast, noise, and anatomical variability. Traditional deep learning models either process frames independently, ignoring temporal information, or rely on memory-based feature propagation, which accumulates error over time. \textbf{Methods:} We propose \textbf{Point-Seg}, a transformer-based segmentation framework that integrates point tracking as a temporal cue to ensure stable and consistent segmentation of myocardium across frames. Our method leverages a point-tracking module trained on a synthetic echocardiography dataset to track key anatomical landmarks across video sequences. These tracked trajectories provide an explicit motion-aware signal that guides segmentation, reducing drift and eliminating the need for memory-based feature accumulation. Additionally, we incorporate a temporal smoothing loss to further enhance temporal consistency across frames. \textbf{Results:} We evaluate our approach on both public and private echocardiography datasets. Experimental results demonstrate that Point-Seg has statistically similar accuracy in terms of Dice to state-of-the-art segmentation models in high quality echo data, while it achieves better segmentation accuracy in lower quality echo with improved temporal stability. Furthermore, Point-Seg has the key advantage of pixel-level myocardium motion information as opposed to other segmentation methods. Such information is essential in the computation of other downstream tasks such as myocardial strain measurement and regional wall motion abnormality detection. \textbf{Conclusion:} Point-Seg demonstrates that point tracking can serve as an effective temporal cue for consistent video segmentation, offering a reliable and generalizable approach for myocardium segmentation in echocardiography videos. The code is available at \href{https://github.com/DeepRCL/PointSeg/}{\texttt{https://github.com/DeepRCL/PointSeg/}}.
}




\maketitle

\section{Introduction}

Echocardiography (echo) is the frontline imaging modality for rapid cardiovascular triage in point of care ultrasound (POCUS) because it is portable, real time, and non invasive~\cite{dlcardiac_review,us_survey}. In this setting, myocardium segmentation provides quantitative information of systolic function at the bedside. Accurate contours enable automated computation of myocardial strain and regional wall motion abnormality, two biomarkers that strongly inform the diagnosis and management of heart failure (HF)~\cite{myoseg_dl}. Yet robust segmentation in POCUS is challenging, where non specialist operators often acquire lower quality echo clips. These realities increase the demand for segmentation methods that are resilient to image quality fluctuations, run in real time, and maintain temporal stability across the cardiac cycle.

Deep learning has advanced cardiac segmentation, but most image based architectures (e.g., UNet~\cite{unet}, SwinUNet~\cite{swinunet}, nnUnet~\cite{isensee2021nnu} operate frame wise and do not enforce temporal coherence, a limitation that becomes acute in echo video analysis. Foundation models such as SAM~\cite{sam} and its medical adaptations (MedSAM~\cite{medsam}, SAMUS~\cite{samus}) generalize across modalities but remain frame independent and often require interactive prompting, impractical in the real time, label scarce POCUS workflow. Video extensions, such as SAM2~\cite{sam2} and MedSAM2~\cite{medsam2}, and memory based echo models like MemSAM~\cite{memsam} improve temporal consistency by propagating features, but they are susceptible to error accumulation when early frames are noisy or off axis, which is common in POCUS.
An alternative line of work enforces temporal consistency using dense optical flow to regularize segmentation across frames~\cite{jafari2018unified,guo2025deformflownet}. However, optical-flow-based approaches rely on brightness constancy and smooth motion assumptions that are frequently violated in echocardiography due to speckle decorrelation, depth-dependent gain, and out-of-plane motion, often leading to temporally unstable or anatomically implausible deformations.

Point tracking offers a motion aware alternative by establishing explicit correspondences through time. Instead of propagating segmentation masks (and compounding errors), tracking learns to follow tissue points across frames, providing a direct temporal signal that is less sensitive to transient artifacts. Long range trajectories have been explored as a self supervisory cue~\cite{lrtl}, but using them to drive myocardium segmentation for bedside EF (Ejection Fraction)/strain estimation remains unexplored. Transformer trackers developed for natural images (TAPTR~\cite{taptrv2}, CoTracker~\cite{cotracker}) demonstrate strong correspondence learning with synthetic pretraining, yet annotated tracking data in echo are limited. EchoTracker~\cite{echotracker} is a step toward echo specific tracking, but small, curated datasets constrain generalization, especially to the image quality variability endemic to POCUS.

We propose a tracking guided myocardium segmentation framework tailored for POCUS HF assessment. Our design prioritizes robustness to image quality variability so that EF/strain can be computed reliably. The model integrates dense point tracking with segmentation to eliminate memory based drift, explicitly models myocardial motion, and retains stability under variations in image quality.

Our key contributions are:
\begin{itemize}[label=$\bullet$]
\item \textbf{Echo-tuned point tracking.} We adapt and fine tune TAPTR for echocardiography using the SynUS synthetic dataset~\cite{synus}, yielding robust, densely sampled myocardial trajectories across frames, even when image quality fluctuates.
\item \textbf{Tracking conditioned segmentation.} We introduce a transformer segmentation head that ingests tracked trajectories as motion priors, enforcing temporal consistency without memory propagation and reducing drift on low quality echo clips.
\end{itemize}

By explicitly modeling myocardial motion and optimizing for robustness to POCUS image quality variation, the proposed method targets what matters most in bedside HF diagnosis: trustworthy, temporally consistent myocardium segmentation that enables actionable measurements.

\section{Method}
We consider an echo cine sequence of $T$ frames, where each frame $I^t \in \mathbb{R}^{H \times W}$ represents an echo image. We track $N_{\text{points}}$ distinct points, including an optional set of manually selected points and a dense grid of regularly spaced points. Each tracked point follows a trajectory $L_i = \{ l_i^t \}_{t=1}^{T}$, where $l_i^t = (x_i^t, y_i^t)$ denotes its position in frame $t$, and a visibility sequence $V_i = \{ v_i^t \}_{t=1}^{T}$, where $v_i^t \in \{0,1\}$ indicates visibility. Using the temporal consistency of these tracked trajectories and spatial image features, we predict segmentation masks $M_k^t \in [0,1]^{H \times W}$ per frame for $k = 1, \dots, K$ classes. We set $K=1$ to segment the myocardium.

The encoder consists of a convolutional neural network (CNN) followed by a Transformer. The CNN extracts multi-scale spatial features, which the Transformer processes to generate patch tokens \( f_{j,s}^t \in \mathbb{R}^d \), where \( j = 1, \dots, N_{\text{patches}} \) indexes tokens, \( s = 1, \dots, S \) denotes scale, and \( t \) represents the frame index. These tokens, capturing contextual and hierarchical features, serve as input to both the point and mask decoder modules.

\subsection{Point Decoder}
The point decoder estimates trajectories and visibilities for the grid of points, producing temporally consistent representations. Given an input sequence, it refines point locations across frames and predicts a binary visibility score indicating whether a point remains visible. The final outputs include point trajectories \( l_i^t \), visibility sequences \( v_i^t \), and point tokens \( p_i^t \), which encode spatiotemporal information about each tracked point. These point tokens are passed to the mask decoder to incorporate motion cues into the segmentation process.


\subsection{Fusion Layers and Mask Decoder}
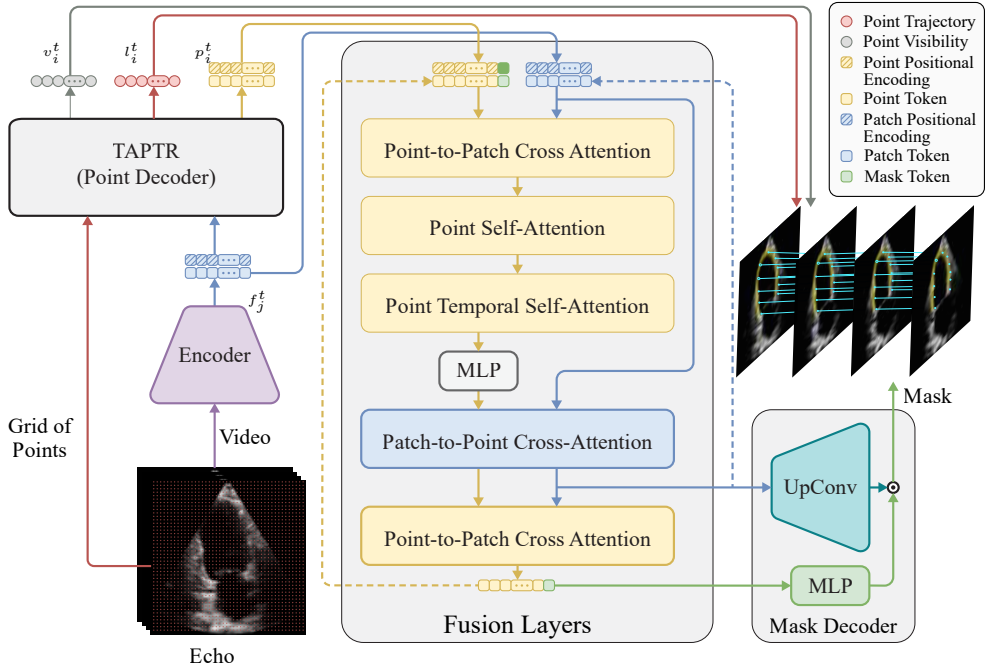
\begin{figure}[t]
    \centering
    \begin{tikzpicture}
    \node[anchor=south west] (img) at (0,0) {\includegraphics[width=.983\textwidth,keepaspectratio]{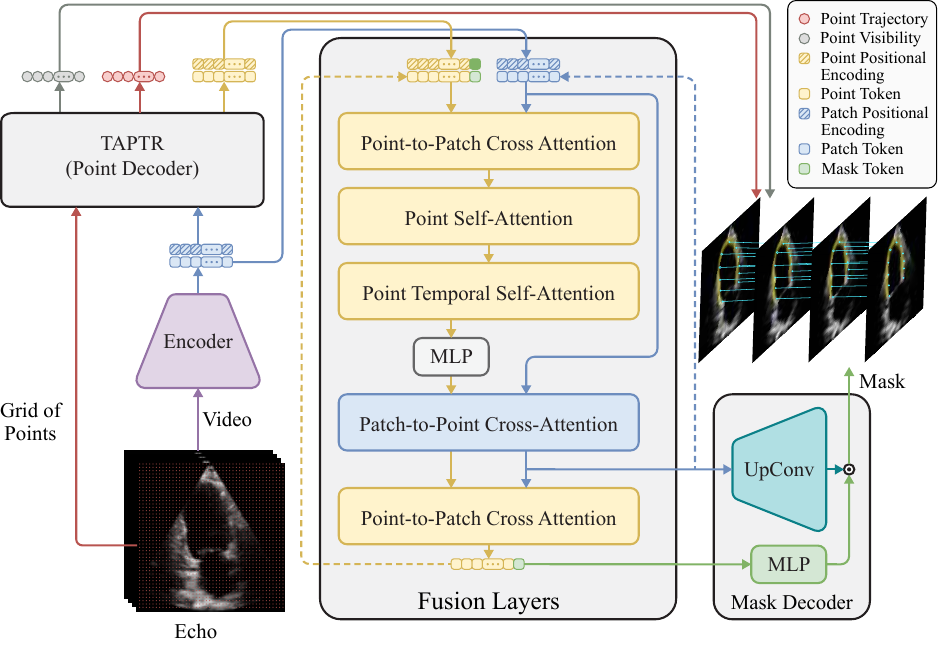}};
    \node at (0.7, 8.3) {\tiny $v_i^t$};
    \node at (1.75, 8.3) {\tiny $l_i^t$};
    \node at (2.7, 8.3) {\tiny $p_i^t$};
    \node at (3.4, 5) {\tiny $f_j^t$};
    \end{tikzpicture}
    \caption{Overview of the proposed model architecture, consisting of a point tracking module and a segmentation module. The point decoder tracks both manually selected and grid-based points, while the mask decoder integrates spatial and temporal cues through attention mechanisms to generate segmentation masks.}
    \label{fig:architecture}
\end{figure}

The fusion layers integrate point and patch tokens to capture spatial relationships within frames and temporal dependencies across frames through attention mechanisms. As shown in Fig.~\ref{fig:architecture}, the refined patch tokens are then processed by the mask decoder to produce accurate segmentation masks.

\subsubsection{Fusion Layers}
The fusion layers employ multiple attention mechanisms to integrate spatial and motion-aware information. Point-to-patch cross-attention allows each point token \( p_i^t \) to attend to patch tokens \( F^t = \{ f_j^t \}_{j=1}^{N_{\text{patches}}} \), incorporating spatial features as \( p_i^t = \text{Attn}(p_i^t, F^t) \). Positional encodings are applied to both points and patches following~\cite{sin_enc}, and the attention mechanism follows~\cite{att_need}. 

Point self-attention enables tracked points to exchange information within a frame, refining their representations based on neighboring points. This is computed as \( p_i^t = \text{Attn}(p_i^t, P^t) \), where \( P^t = \{ p_i^t \}_{i=1}^{N_{\text{points}}} \). Point temporal self-attention models motion by allowing each point token to attend to its trajectory across frames, enhancing consistency over time. The updated tokens are given by \( p_i^t = \text{Attn}(p_i^t, P_i) \), where \( P_i = \{ p_i^t \}_{t=1}^{T} \).

Patch-to-point cross-attention integrates motion-aware information into spatial representations, ensuring segmentation consistency while preserving details. Each patch token \( f_j^t \) attends to point tokens as \( f_j^t = \text{Attn}(f_j^t, P^t) \), where \( P^t = \{ p_i^t \}_{i=1}^{N_{\text{points}}} \). The resulting point- and motion-aware patch representations are then passed to the mask decoder for pixel-level reconstruction.


\subsubsection{Mask Decoder}
The mask decoder transforms the fused patch representations into dense segmentation masks. We introduce $K$ learnable mask tokens \( m_k \in \mathbb{R}^d \) following~\cite{sam}, which represent segmentation classes. These tokens are concatenated with point tokens and processed through attention layers to incorporate spatial and motion-aware features.

Subsequently, mask tokens are then transformed via a multi-layer perceptron (MLP) to align with the upscaled patch features. Patch tokens undergo progressive upsampling using transposed convolutions, refining multi-scale features until reaching the input resolution. The segmentation masks are obtained via an inner product between mask tokens and upscaled patch tokens:
\begin{equation}
    M_k^t = \sigma \left( \text{MLP}(m_k^t) \cdot F^t \right),
\end{equation}
where \( M_k^t \) is the predicted mask for class \( k \) at frame \( t \), and \( \sigma(\cdot) \) denotes the sigmoid activation. The fusion layers comprise \( N_{\text{FL}} \) transformer blocks that iteratively refine patch, point, and mask tokens, ensuring stable and temporally consistent segmentation.

\subsection{Learning Objectives}
We optimize the model for accurate segmentation and temporally stable tracking. Training is performed in two stages: we first train the point decoder, then freeze it and train the mask decoder. The point decoder is trained with an $\ell_1$ loss for trajectory estimation and a cross-entropy loss for visibility prediction. The total tracking loss is:
\begin{equation}
    \mathcal{L}_{\text{points}} = w_V \mathcal{L}_\text{CE} (V^{N_{\text{pd}}}, \tilde{V}) + \sum_{n=1}^{N_{\text{pd}}} w_L^n \|L^n - \tilde{L}\|_1.
\end{equation}
Here, \( L^n \) and \( \tilde{L} \) are the predicted and ground-truth trajectories, while \( V^{N_{\text{pd}}} \) and \( \tilde{V} \) are the predicted and ground-truth visibility sequences from the final layer.

After training the point decoder, we freeze it and train the mask decoder using the Dice loss~\cite{dice}. The total segmentation loss is:
\begin{equation}
    \mathcal{L}_\text{Mask} = \sum_{n=1}^{N_{\text{FL}}} w_M^n \mathcal{L}_\text{Dice}(M_n, \tilde{M}),
\end{equation}
where \( M_n \) are the predicted masks at layer \( n \), \( \tilde{M} \) are the ground truth masks.

To enforce temporal consistency in segmentation, we extend the temporal smoothing loss from~\cite{lrtl} by enforcing pairwise similarity between tracked mask values across frames. Given the predicted masks \( M_n^t \), bilinear sampling is applied at tracked points \( M_n^t(L^t) \), and the loss is:
\begin{equation}
    \mathcal{L}_\text{Temp} = \sum_{n=1}^{N_{\text{FL}}} w_T^n \sum_{t_1 < t_2} \left\| M_n^{t_1}(L^{t_1}) - M_n^{t_2}(L^{t_2}) \right\|_2^2.
\end{equation}

\section{Experiments and Results}
\subsection{Datasets}

We train and evaluate our model using three echocardiography datasets. The CAMUS dataset~\cite{camus} includes 500 patients with annotated apical two- and four-chamber (AP2, AP4) sequences. It is split by patient into 350/50/100 for training, validation, and testing, ensuring both views from each patient remain in the same split. The test set contains 322 poor-quality and 3,545 medium/good-quality frames.

For point-tracking supervision, we use the synthetic SynUS dataset~\cite{synus}, which provides dense myocardial motion and visibility annotations generated from 50 high-quality CAMUS cases, following the same split protocol.

For clinical evaluation, we also train on a private dataset comprising 1,305 training, 181 validation, and 367 test videos with annotations available only at end-diastolic (ED) and end-systolic (ES) frames. The test set includes 40 poor-quality and 694 medium/good-quality frames, with quality assessed using an automatic echo quality estimation model~\cite{liao2019modelling}.

\begin{table}[t]
\caption{\textbf{Segmentation performance} on the \textbf{CAMUS}~\cite{camus} \textbf{dataset}. 
Models are trained and evaluated on CAMUS, with results reported for the Myocardium (Myo) across different image quality groups. 
Column \textbf{P} indicates whether the model requires a manual \emph{Prompt}. 
\textbf{mDice}: mean Dice coefficient (0–100); \textbf{HD95}: 95th percentile Hausdorff distance (in millimeters). For MedSAM2, “click” and “bbox” denote the type of prompt used, corresponding to a point and a bounding box, respectively. The final row reports paired t-test p-values comparing our model with the best-performing baseline}
\label{tab:segmentation_results}
\centering
\begin{tabular*}{\textwidth}{@{\extracolsep\fill}lccccc}
\toprule
\multirow{3}{*}{\textbf{Method}} 
 & \multirow{3}{*}{\textbf{P}} 
 & \multicolumn{4}{@{}c@{}}{\textbf{CAMUS (test)}~\cite{camus}} \\ 
\cmidrule(lr){3-6}
 &  & \multicolumn{2}{c}{\textbf{Good/Medium Quality}} & \multicolumn{2}{c}{\textbf{Poor Quality}} \\ 
\cmidrule(lr){3-4} \cmidrule(lr){5-6}
 &  & mDice$\uparrow$ & HD95$\downarrow$ & mDice$\uparrow$ & HD95$\downarrow$ \\ 
\midrule
UNet~\cite{unet} & \xmark & 83.8(6.1) & 5.2(3.9) & 84.0(5.1) & 6.0(4.4) \\
SwinUNet~\cite{swinunet} & \xmark & 84.3(5.7) & 5.0(2.2) & 84.4(5.4) & 5.6(2.6) \\
nnUNet~\cite{isensee2021nnu} & \xmark & \underline{85.7(5.5)} & \underline{4.5(2.4)} & \underline{85.4(5.8)} & \underline{4.8(2.2)} \\
\midrule
DeformFlowNet~\cite{guo2025deformflownet} & \xmark & 84.2(6.1) & 7.5(8.0) & 82.0(7.0) & 10.1(8.8) \\
MedSAM2~(click)~\cite{medsam2} & \cmark & 77.7(13.3) & 13.6(16.5) & 74.0(15.1) & 16.7(16.8) \\
MedSAM2~(bbox)~\cite{medsam2} & \cmark & 85.2(5.0) & 4.6(1.6) & 84.2(5.2) & 5.5(1.8) \\
MemSAM~\cite{memsam} & \cmark & 85.4(5.7) & 5.3(4.8) & 80.8(10.9) & 9.5(9.5) \\
\textbf{PointSeg (ours)} & \xmark & \textbf{85.8(5.4)} & \textbf{4.4(1.6)} & \textbf{86.7(5.4)} & \textbf{4.7(2.3)} \\
\bottomrule
p-value & & $p\geq0.05$ & $p\geq0.05$ & $p<0.05$ & $p\geq0.05$ \\
\bottomrule
\end{tabular*}
\end{table}

\subsection{Implementation Details}


We use TAPTR~\cite{taptrv2} as our point tracking model and fine-tune it on the SynUS dataset~\cite{synus}, utilizing 800 tracked sequences and 10 consecutive frames per recording.
For the segmentation pipeline, we employ an encoder based on a ResNet50~\cite{resnet50} backbone with two transformer encoder layers using deformable attention~\cite{DETR}. The mask decoder consists of three layers ($N_{\text{FL}}=3$) with layer-wise weights of 0.25, 0.5, and 1.0. Training is performed on two V100 GPUs for 200 epochs using AdamW~\cite{adamw} (learning rate $2\times10^{-4}$, weight decay $1\times10^{-4}$), a batch size of one, gradient accumulation every four steps, and input frames resized to $256\times256$ for fair comparison with prior baselines, although this is not a limitation of the proposed method. Dice and temporal smoothing loss weights are set to 1.0 and 100, respectively. At inference, the processing speed is approximately 32.8 FPS for this setup, which depends on the input resolution and the number of tracked points; the reported speed corresponds to the configuration used in our experiments.

\subsection{Comparison with State-of-the-Art Methods}
For a fair comparison, all baseline models were trained or fine-tuned on the corresponding training datasets following the same protocol as our method. For models with publicly available checkpoints, fine-tuning was performed after loading the pretrained weights, provided that the CAMUS split configuration did not introduce overlap or data leakage.

\subsubsection{Quantitative Results}

Table~\ref{tab:segmentation_results} summarizes results on CAMUS, and Table~\ref{tab:rwma_results} shows results on the private echocardiography dataset, where videos were categorized into good/medium and poor-quality groups. In both datasets, the p-values for good/medium quality indicate no statistically significant difference among models, suggesting that most segmentation methods perform similarly when the image quality is sufficient. However, for poor-quality cases, the performance differences become statistically significant, revealing that conventional models degrade under challenging imaging conditions. 
In this setting, conventional models that process frames independently or lack explicit temporal modeling exhibit notable performance degradation. 
Similarly, optical-flow-based temporal regularization in DeformFlowNet\cite{guo2025deformflownet} shows inferior performance compared to PointSeg, specially in poor-quality videos, likely due to the sensitivity of dense optical flow to speckle decorrelation and out-of-plane motion in echocardiography, which leads to unstable temporal constraints.
Prompt-based methods such as MedSAM2 and MemSAM show strong performance on the private dataset, particularly when bounding-box prompts are provided. This behavior is expected given the characteristics of the dataset: each video contains only two frames (ED and ES), meaning that a prompt provided on the first frame directly constrains half of the sequence, substantially simplifying the segmentation task.

In contrast, PointSeg does not rely on manual prompts and instead uses automatically generated point trajectories to enforce temporal consistency. Unlike MedSAM2 and MemSAM, which leverage pretrained SAM-based foundation models and user-provided prompts, PointSeg is trained only on synthetic tracking data and CAMUS (and the private dataset for baseline consistency). Despite these differences, PointSeg remains robust in poor-quality videos, achieving reliable segmentation and boundary precision without user interaction.


\begin{figure}[t]
    \centering
    \begin{tikzpicture}
    \node[anchor=south west] (img) at (0,0)
    {\includegraphics[width=.93\textwidth,keepaspectratio]{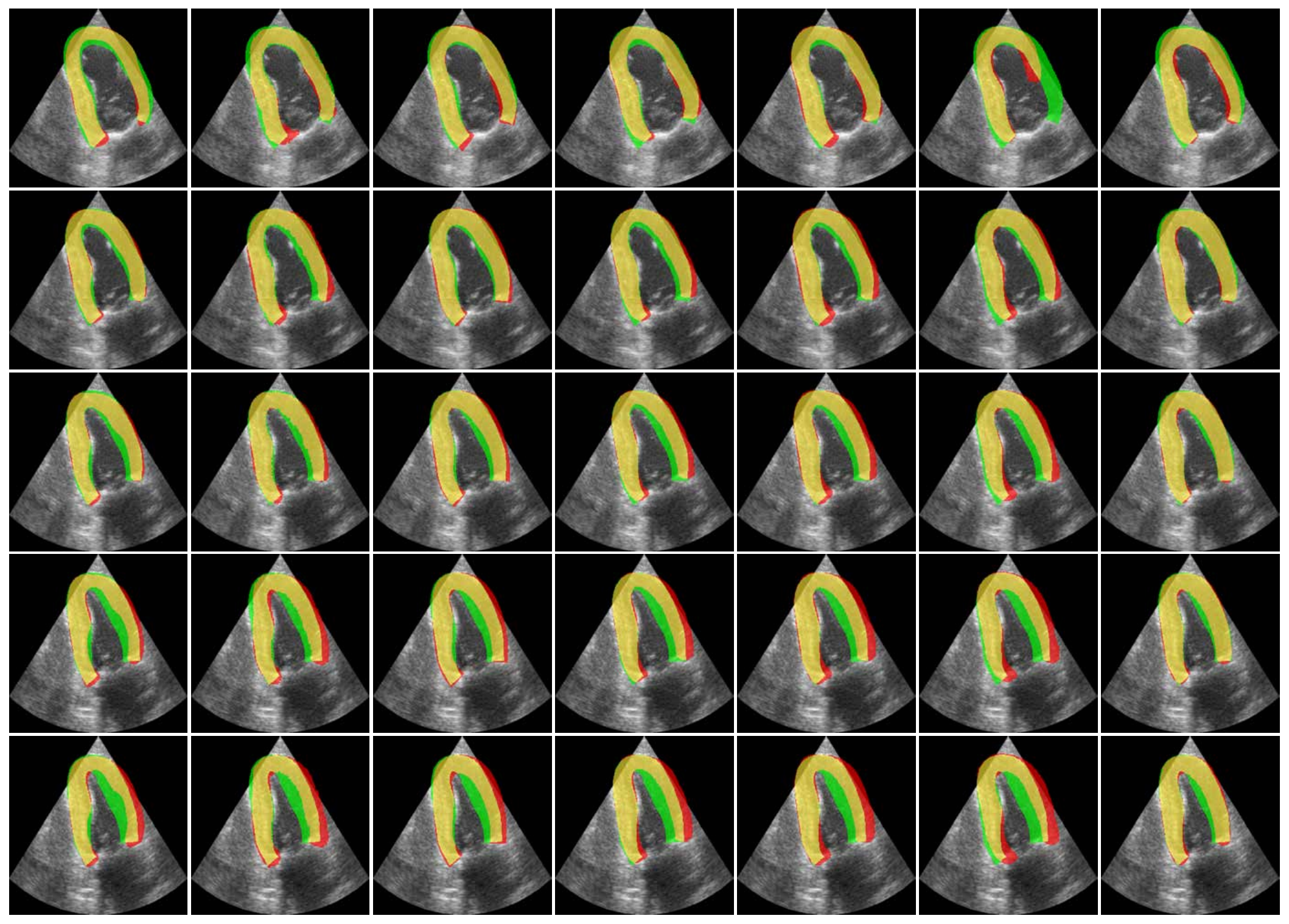}};
    \node[scale=0.7] at (1.1, 0) {\tiny UNet};
    \node[scale=0.7] at (2.8, 0) {\tiny SwinUNet};
    \node[scale=0.7] at (4.5, 0) {\tiny nnUNet};
    \node[scale=0.7] at (6.2, 0) {\tiny MedSAM2};
    \node[scale=0.7] at (6.2, -0.3) {\tiny (click)};
    \node[scale=0.7] at (7.9, 0) {\tiny MedSAM2};
    \node[scale=0.7] at (7.9, -0.3) {\tiny (bbox)};
    \node[scale=0.7] at (9.65, 0) {\tiny MemSAM};
    \node[scale=0.7] at (11.4, 0) {\tiny PointSeg};
    \node[scale=0.7] at (11.4, -0.3) {\tiny (ours)};

    \node[scale=0.7][rotate=90] at (-0.09, 1.05) {\tiny t = 14};
    \node[scale=0.7][rotate=90] at (-0.09, 2.85) {\tiny t = 12};
    \node[scale=0.7][rotate=90] at (-0.09, 4.6) {\tiny t = 10};
    \node[scale=0.7][rotate=90] at (-0.09, 6.3) {\tiny t = 7};
    \node[scale=0.7][rotate=90] at (-0.09, 8.0) {\tiny t = 1};
    \end{tikzpicture}
    \caption{Qualitative comparison of segmentation results for one echo from CAMUS dataset. \textbf{Red} represents \textbf{predictions}, \textbf{green} denotes \textbf{ground truth}, and \textbf{yellow} indicates \textbf{overlap}. Our method shows no flickering compared to image-based models and less drift than memory-based approaches.}
    \label{fig:comparison}
\end{figure}
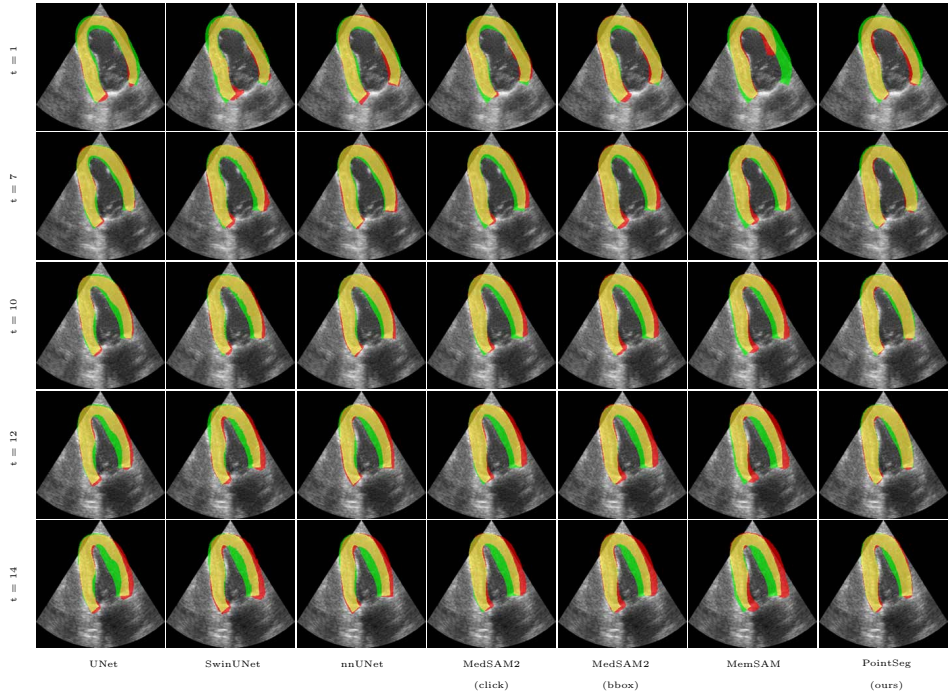

\begin{table}[t]
\caption{\textbf{Segmentation performance} on the \textbf{Private dataset}. 
Models are trained and evaluated on CAMUS and Private data, with results reported for the Myocardium (Myo) across different image quality groups.
\textbf{mDice}: mean Dice coefficient (0–100); \textbf{HD95}: 95th percentile Hausdorff distance (in millimeters). The final row reports paired t-test p-values comparing our model with the best-performing baseline.}
\label{tab:rwma_results}
\centering
\begin{tabular*}{\textwidth}{@{\extracolsep\fill}lccccc}
\toprule
\multirow{3}{*}{\textbf{Method}} 
 & \multicolumn{4}{@{}c@{}}{\textbf{Private Data (test)}~\cite{camus}} \\ 
\cmidrule(lr){2-5}
 & \multicolumn{2}{c}{\textbf{Good/Medium Quality}} & \multicolumn{2}{c}{\textbf{Poor Quality}} \\ 
\cmidrule(lr){2-3} \cmidrule(lr){4-5}
 & mDice$\uparrow$ & HD95$\downarrow$ & mDice$\uparrow$ & HD95$\downarrow$ \\ 
\midrule
UNet~\cite{unet} & 69.3(10.6) & 6.1(5.1) & 64.8(10.6) & 9.7(7.9) \\
SwinUNet~\cite{swinunet} & 68.3(10.0) & 6.5(4.1) & 62.3(12.3) & 9.0(5.3) \\
nnUNet~\cite{isensee2021nnu} & 69.7(11.1) & \underline{5.6(3.1)} & 61.7(11.8) & 10.0(8.9) \\
\midrule
DeformFlowNet~\cite{guo2025deformflownet} & 67.4(10.7) & 12.7(14.1) & 58.9(10.4) & 17.3(14.2) \\
MedSAM2~(click)~\cite{medsam2} & 63.6(15.5) & 9.6(10.4) & 56.2(14.1) & 13.9(9.7) \\
MedSAM2~(bbox)~\cite{medsam2} & \textbf{72.4(9.9)} & \textbf{5.0(1.8)} & \textbf{68.1(10.5)} & \textbf{6.5(2.7)} \\
MemSAM & \underline{69.8(10.9)} & 6.7(6.5) & 65.4(12.4) & 14.8(14.4) \\
\textbf{PointSeg (ours)} & \underline{69.8(10.0)} & 5.9(2.1) & \underline{66.6(10.1)} & \underline{7.5(2.8)} \\
\bottomrule
p-value & $p<0.05$ & $p<0.05$ & \textbf{$p\geq0.05$} & \textbf{$p < 0.05$} \\
\bottomrule
\end{tabular*}
\end{table}

\subsubsection{Qualitative Results}
Figure~\ref{fig:comparison} presents qualitative comparisons of segmentation results across different models. UNet~\cite{unet}, SwinUNet~\cite{swinunet}, and nnUNet~\cite{isensee2021nnu} show increasing segmentation discrepancies over time, where the predicted masks gradually deviate from the ground truth as the frame index $t$ increases. Similarly, memory-based models such as MemSAM~\cite{memsam} and MedSAM2~\cite{medsam2} exhibit noticeable segmentation drift, especially in low-quality frames. In contrast, our proposed PointSeg maintains stable and temporally coherent predictions throughout the cardiac cycle, closely following the true anatomical trajectories. The consistent boundary alignment over time highlights the effectiveness of incorporating point tracking as an explicit temporal cue for segmentation stability.
\subsection{Ablation Study}
We evaluate the impact of point trajectories and individual transformer block components on segmentation performance. Specifically, we analyze the contributions of point-to-patch cross-attention, the MLP layer, point self-attention, and temporal self-attention. All experiments are conducted on CAMUS~\cite{camus}, using all frames for consistency. Table~\ref{tab:components} shows that progressively adding these components improves segmentation, with temporal self-attention playing a key role in refining features across frames. The results also indicate that incorporating point trajectories enhances stability, highlighting the importance of temporal information for robust segmentation. The reported p-values further show that the performance differences between each ablation variant and the final model are statistically significant, confirming the contribution of each component.

\begin{table}[t]
\centering
\caption{Ablation study on the impact of point trajectories and mask decoder components in segmentation performance. Starting from the left, the columns indicate the presence or absence of point trajectories, point-to-patch cross-attention, MLP layer, point self-attention, and point temporal self-attention, respectively. The p-value column reports statistical significance with respect to the final model (last row).}
\label{tab:components}

\begin{tabular}{c|c|c|c|c||c|c||c}
\toprule
Points & Point-to-Patch CA & MLP & Point SA & Point TSA & mDice$\uparrow$ & HD95$\downarrow$ & p-value \\
\midrule
\xmark & \cmark & \cmark & \cmark & \cmark & 84.36 & 5.12 & $p<0.05$ \\
\midrule
\cmark & \cmark & \xmark & \xmark & \xmark & 84.84 & 5.05 & $p<0.05$ \\
\cmark & \cmark & \cmark & \xmark & \xmark & 85.05 & 4.66 & $p<0.05$ \\
\cmark & \cmark & \cmark & \cmark & \xmark & 84.40 & 4.81 & $p<0.05$ \\
\cmark & \cmark & \cmark & \cmark & \cmark & \textbf{85.85} & \textbf{4.39} & - \\
\bottomrule
\end{tabular}
\end{table}

To validate our choice of trajectory source, we also compare our tracker against state-of-the-art methods. As shown in Table~\ref{tab:tracking_results}, our approach surpasses EchoTracker~\cite{echotracker}, a leading method for myocardial tracking in echocardiography videos, as well as its variants, achieving stronger motion estimation and occlusion robustness, which in turn benefits segmentation performance.

\begin{table}[t]
\caption{\textbf{Point tracking performance} on the \textbf{synthetic} dataset. 
The results are evaluated using Average Jaccard (AJ), Average Points Within Range ($<\delta_{avg}$), and Occlusion Accuracy (OA). The final row reports paired t-test p-values comparing our model with the best-performing baseline for each metric.}
\label{tab:tracking_results}
\centering
\begin{tabular*}{\textwidth}{@{\extracolsep\fill}llccc}
\toprule
\multirow{2}{*}{\textbf{Method}} & \multirow{2}{*}{\textbf{Training Data}} & 
\multicolumn{3}{@{}c@{}}{\textbf{SynUS (test)}~\cite{synus}} \\
\cmidrule(lr){3-5}
 & & AJ$\uparrow$ & $<\delta_{avg}$$\uparrow$ & OA$\uparrow$ \\ 
\midrule
EchoTracker~\cite{echotracker} & Private Data & 0.764 & 0.832 & - \\
EchoTracker Fine-tuned~\cite{echotracker} & Private Data + SynUS~\cite{synus} & \underline{0.871} & \underline{0.918} & - \\
CoTracker~\cite{cotracker} & Kubric~\cite{kubric} & 0.700 & 0.887 & 0.836 \\
CoTracker Fine-tuned~\cite{cotracker} & Kubric~\cite{kubric} + SynUS & 0.832 &  0.898 & 0.985 \\
TAPTR~\cite{taptrv2} & Kubric~\cite{kubric} & 0.695 & 0.834 & \underline{0.894} \\
\textbf{TAPTR Fine-tuned (ours)} & Kubric~\cite{kubric} + SynUS & \textbf{0.888} & \textbf{0.934} & \textbf{0.992} \\
\midrule
 p-value &
& $p{<}0.05$ & $p{<}0.05$ & $p{<}0.05$ \\
\bottomrule
\end{tabular*}
\end{table}

\section{Conclusion and Future Work}

We introduced a transformer-based model that leverages point tracking as a temporal cue for myocardium segmentation in echocardiography videos. By incorporating motion-aware point trajectories, our approach enhances temporal consistency and produces smoother, more anatomically coherent segmentation across frames. Experiments show that our model performs on par with state-of-the-art methods on high-quality echocardiography samples, but achieves markedly better performance on low-quality data, while requiring no manual prompts and simultaneously predicting point trajectories that guide temporally consistent segmentations.

In future work, we aim to extend this framework toward myocardial function analysis by using the tracked trajectories for quantitative strain measurement. While the proposed model is designed for echocardiography, its applicability to other ultrasound modalities would require separate evaluation and validation. We also plan to implement online tracking and segmentation, as the current version operates in an offline setting, although the proposed architecture does not impose any inherent limitation on online inference.

\bibliography{sn-bibliography}

\end{document}